\documentclass[10pt,twocolumn,letterpaper]{article}

\usepackage[pagenumbers]{cvpr} % To force page numbers, e.g. for an arXiv version

\usepackage[pagebackref,breaklinks,colorlinks]{hyperref}
\usepackage{tcolorbox}
\usepackage{multirow}
\usepackage{pifont}
\usepackage{hyperref}

\title{VTimeLLM: Empower LLM to Grasp Video Moments}

\author{Bin Huang, Xin Wang\thanks{Corresponding Authors.}, Hong Chen, Zihan Song, Wenwu Zhu\footnotemark[1]\\
Tsinghua University\\
{\tt\small \{huangb23, h-chen20, songzh23\}@mails.tsinghua.edu.cn} \\
{\tt\small \{xin\_wang,wwzhu\}@tsinghua.edu.cn}
}

\begin{document}
\maketitle
\begin{abstract}
Large language models (LLMs) have shown remarkable text understanding capabilities, which have been extended as Video LLMs to handle video data for comprehending visual details. 
However, existing Video LLMs can only provide a coarse description of the entire video, failing to capture the precise start and end time boundary of specific events. 
In this paper, we solve this issue via proposing VTimeLLM, a novel Video LLM designed for fine-grained video moment understanding and reasoning with respect to time boundary. 
Specifically, our VTimeLLM adopts a boundary-aware three-stage training strategy, which respectively utilizes image-text pairs for feature alignment, multiple-event videos to increase temporal-boundary awareness, and high-quality video-instruction tuning to further improve temporal understanding ability as well as align with human intents. Extensive experiments demonstrate that in fine-grained time-related comprehension tasks for videos such as Temporal Video Grounding and Dense Video Captioning, VTimeLLM significantly outperforms existing Video LLMs. Besides, benefits from the fine-grained temporal understanding of the videos further enable VTimeLLM to beat existing Video LLMs in video dialogue benchmark, showing its superior cross-modal understanding and reasoning abilities. \footnote{Our project page is at \url{https://github.com/huangb23/VTimeLLM}}
\end{abstract}    
\section{Introduction}
\label{sec:intro}

\begin{figure}[htbp]
  \centering
  \includegraphics[width=.9\columnwidth]{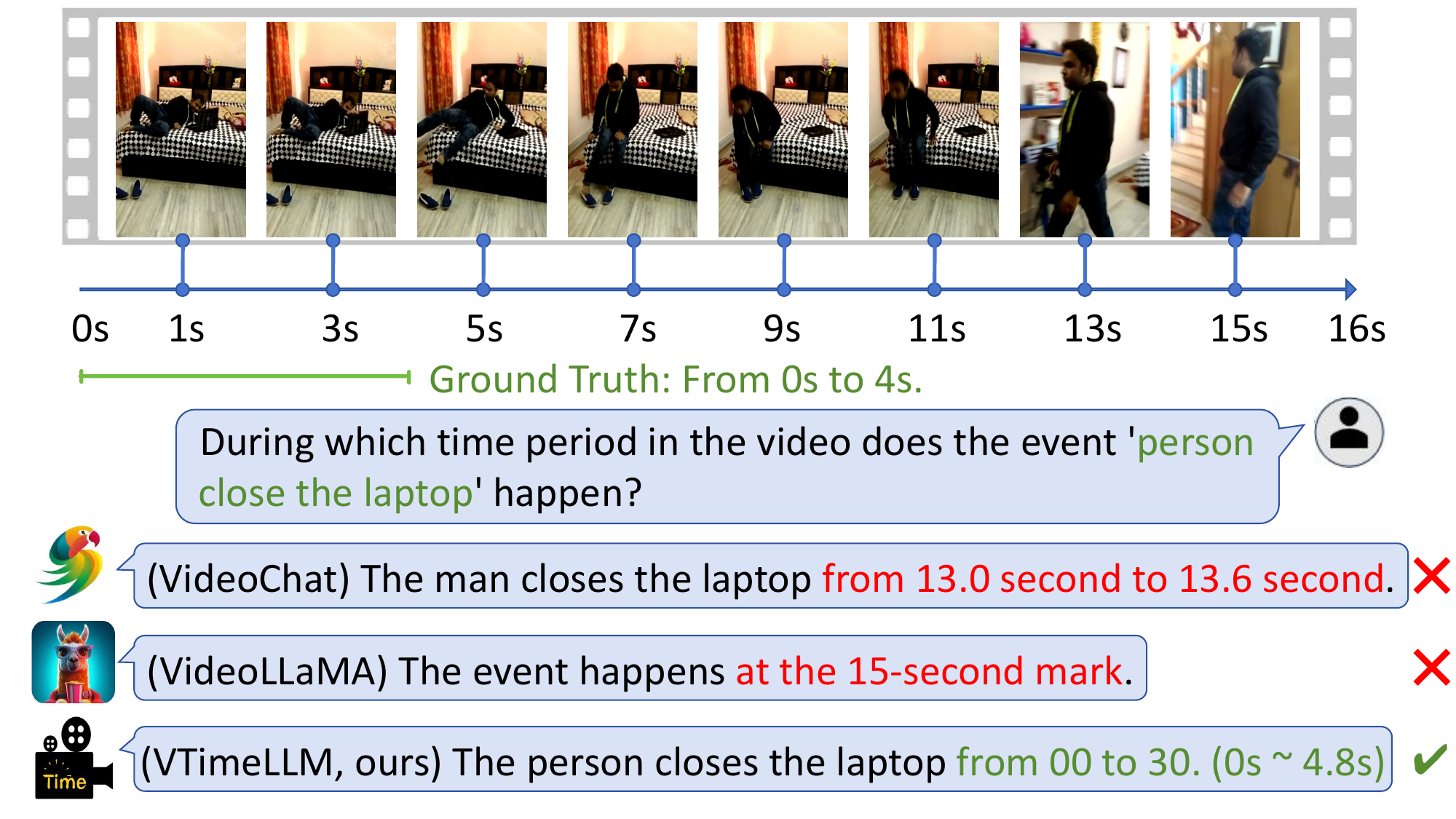}
  \caption{Existing Video LLMs, such as VideoChat and VideoLLaMA, exhibit a deficiency in boundary awareness, leading to challenges in accurately capturing the precise timestamps of specific events.}
  \label{fig:intro}
\end{figure}

Large language models (LLMs) have garnered significant attention due to their exceptional capabilities in text understanding and generation~\cite{zeng2022glm, touvron2023llama2}. However, harnessing the potential of LLMs for understanding and reasoning over multimodal data, especially videos, still remains a substantial challenge. This is because analyzing videos requires deep understanding of both visual details and temporal dynamics for models. 
Several preliminary attempts~\cite{li2023videochat, maaz2023videochatgpt, zhang2023videollama} utilize LLMs for video understanding. Nevertheless, these works predominantly focus on generation of generic video captions and can merely offer surface-level summaries of the content, thus failing to capture the relationships between specific moment boundaries and the bounded events, as shown in Figure~\ref{fig:intro}.

To tackle the problem, in this paper, we investigate improving the boundary-aware ability of Video LLM, which faces the following two challenges. 
\begin{enumerate}
    \item There is a scarcity of large-scale video datasets with accurate boundary annotations to train the Video LLM for temporal alignment.
    \item It is non-trivial to design effective temporal-related video tasks for training LLM to understand the content of multiple moments within videos.
\end{enumerate}

To address these challenges, we propose VTimeLLM, a novel Video LLM that can perceive fine-grained segments in videos with better temporal reasoning ability. 
VTimeLLM consists of i) a visual encoder and a visual adapter to process the input video, and ii) a tailored LLM to understand both text and video content, which is trained via a novel boundary-aware three-stage training strategy. 
Specifically, visual features are aligned with LLM's semantic space through image-text training in the first stage. In the second stage, we then design the single-turn and multi-turn related question answering (QA) tasks to 
endue VTimeLLM with the awareness of time boundaries and the ability to understand the corresponding events bounded within the boundaries. We employ a large-scale video-text dataset containing multiple segments together with 
their roughly annotated labels for training VTimeLLM with the QA tasks. Finally, in the third stage, we further create a high-quality dialogue dataset for instruction tuning, which simultaneously aligns VTimeLLM with human intention and enables VTimeLLM to conduct temporal understanding for video segments more precisely. 
Extensive experiments show that VTimeLLM significantly outperforms existing Video LLMs in time-related video understanding tasks, such as Temporal Video Grounding and Dense Video Captioning. 
In addition, benefiting from the fine-grained temporal understanding of videos, VTimeLLM is able to beat existing Video LLMs in video dialogue benchmark, demonstrating its superiority in cross-modal understanding and reasoning for videos. Our contributions in this paper are listed as follows,

\begin{itemize}
\item We propose VTimeLLM, the first boundary-aware Video LLM, to the best of our knowledge.

\item We propose the boundary-aware three-stage training strategy, which consecutively leverages i) large-scale image-text data for feature alignment, ii) large-scale multi-event video-text data together with the temporal-related single-turn and multi-turn QA to enhance the awareness of time boundary, and iii) instruction tuning on the high-quality dialog dataset for better temporal reasoning ability.

\item We conduct extensive experiments to demonstrate that the proposed VTimeLLM significantly outperforms existing Video LLMs in various fine-grained temporal-related video tasks, showing its superior ability for video understanding and reasoning.
\end{itemize}

\section{Related Works}

\subsection{Multimodal Large Language Model}

\paragraph{Image LLMs} To enable Large Language Models (LLMs) to comprehend visual information, significant efforts have been made to align visual and linguistic modalities. BLIP-2~\cite{li2023blip2} introduced the concept of Q-Former, utilizing learnable query vectors to extract visual features from frozen image encoders. MiniGPT-4~\cite{zhu2023minigpt} demonstrated that further fine-tuning with detailed image descriptions significantly enhances its usability. LLAVA~\cite{liu2023llava} explored diverse multi-modal instruction-following data, including conversations, detailed descriptions, and complex reasoning, aiming to construct a general-purpose visual assistant. Recent endeavors, such as Kosmos-2~\cite{peng2023kosmos2} and VisionLLM~\cite{wang2023visionllm}, delved into more detailed aspects of image comprehension, including referring and grounding, significantly enhancing the capability to describe intricate image details.

\paragraph{Video LLMs} Driven by the success of Image LLM, researchers have naturally extended their focus from single-frame images to multi-frame videos, leading to the emergence of Video-compatible LLMs like VideoChat~\cite{li2023videochat}, Video-LLaMA~\cite{zhang2023videollama}, and Video-ChatGPT~\cite{maaz2023videochatgpt}. These models employ a two-stage training strategy. In the first stage, large-scale datasets align video features with the feature space of LLMs. In the second stage, a limited amount of GPT-annotated or human-annotated datasets are used for instruction tuning. While these models exhibit impressive overall video comprehension, their abilities to describe specific video segments and perform temporal reasoning remain limited. The limitation arises mainly due to the nature of datasets used in the first training stage, such as WebVid~\cite{bain2021webvid}, which usually consist of one-event videos and noisy textual annotations. Moreover, the scarcity of high-quality, temporally annotated data in the second stage poses a challenge for models to conduct temporal reasoning. To bridge this gap, our approach, VTimeLLM, introduces a boundary perception stage between these two stages. This stage enables the model to precisely locate events within videos and describe multiple distinct events accurately, empowering our model to grasp fine-grained details of video moments.

\subsection{Fine-Grained Video Understanding}

Fine-grained video understanding, the ability to precisely locate and comprehend specific events within a video, is a crucial challenge for video analysis. When integrated with natural language, there are two primary tasks: Temporal Video Grounding~\cite{gao2017tall_grounding9, anne2017localizing_grounding10} and Dense Video Captioning~\cite{krishna2017dense_dense45, wang2018bidirectional_dense96}.
\paragraph{Temporal Video Grounding} Temporal Video Grounding aims to identify corresponding video segments for given textual inputs. Traditional approaches can be categorized into two types: proposal-based~\cite{zhang2019man_grounding16, yuan2019semantic_grounding17, chen2018temporally_grounding14} and proposal-free methods~\cite{yuan2019find_grounding19, ghosh2019excl_grounding20, zhang2020span_grounding23}. Proposal-based techniques generate candidate proposals before ranking them based on relevance. In contrast, proposal-free methods directly predict the start and end boundaries of the target moment.
\paragraph{Dense Video Captioning} Dense Video Captioning is a more intricate task, demanding both temporal localization and captioning for all events within an untrimmed video. Earlier methods~\cite{duan2018weakly, iashin2020better_dense36, krishna2017dense_dense45} employed a two-stage process involving temporal localization followed by event captioning. Recent developments~\cite{wang2018bidirectional_dense96, zhou2018end_dense127, yang2023vid2seq} in this field have witnessed a shift towards joint training of captioning and localization modules. For instance, Vid2Seq~\cite{yang2023vid2seq}, enhances a language model by incorporating specific time tokens, enabling the model to generate event boundaries and textual descriptions within the unified output sequence.

Both these two tasks share a fundamental requirement: the alignment of video segments with semantic context. Leveraging the power of LLM with the help of our training strategy, our VTimeLLM model unifies these tasks and has demonstrated remarkable effectiveness. Concurrently, VTimeLLM enables natural language interaction with humans, establishing itself as an excellent assistant for comprehending video content.
\section{VTimeLLM: Being Aware of Time Boundaries in Videos}

\begin{figure*}[!htbp]
  \centering
  \includegraphics[height=8cm, width=0.9\textwidth]{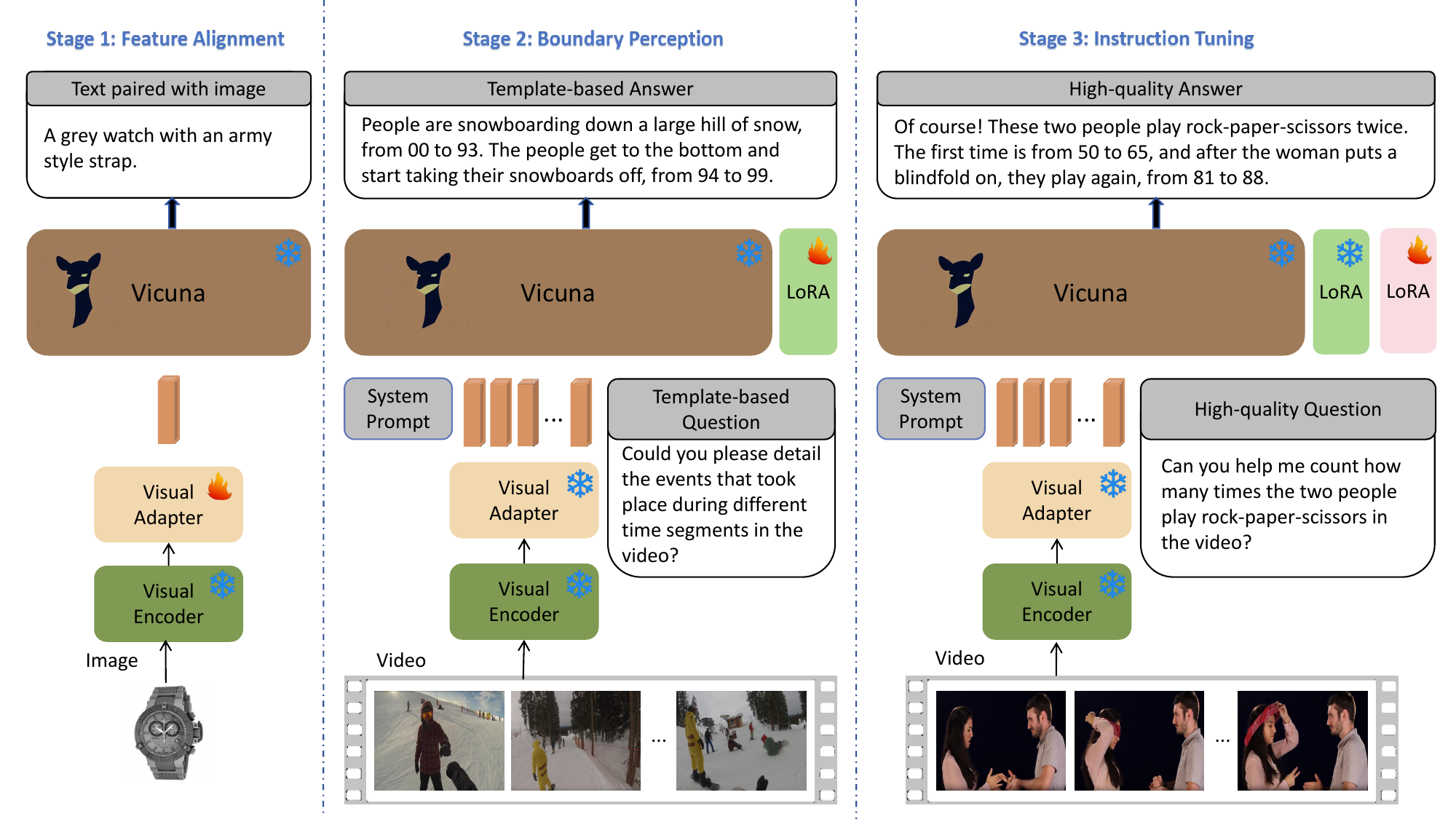}
  \caption{Our boundary-aware three-stage training framework. In the first stage, visual features are aligned with LLM's semantic space through image-text training. In the second stage, we transform a large-scale multi-event dataset into a QA format based on templates, training VTimeLLM to possess temporal boundary awareness and understand events within the boundaries. In the third stage, we create a high-quality dialogue dataset for instruction tuning, which aligns VTimeLLM with human intent and also enables more precise video temporal understanding.}
  \label{fig:framework}
\end{figure*}

In this section, we introduce VTimeLLM,  which is designed to grasp precise video moments for LLMs. We first provide a detailed description of the model architecture, and then our innovative boundary-aware three-stage training framework, as shown in Figure~\ref{fig:framework}.

\subsection{Architecture}

To enable the LLM to comprehend videos, our VTimeLLM model incorporates two additional modules within LLM, i.e., the visual encoder and the visual adapter, which transform the visual information into text space.

\paragraph{Visual Encoder} Our VTimeLLM model utilizes a frozen CLIP ViT-L/14~\cite{clip} as the visual encoder, and for simplicity we denote it as ViT. Given a video $V\in \mathbb{R}^{T \times H \times W \times C}$ with $T$ frames, we uniformly sample $N=100$ frames, represented as $\tilde{V} \in \mathbb{R}^{N \times H \times W \times C}$, where $\tilde{V}_1 = V_1 $ and  $\tilde{V}_N = V_T$. Each frame $\tilde{V}_i$ is independently processed through the visual encoder:
\begin{equation}
    \{v_i^{cls}, v_i^{1}, v_i^{2}, ..., v_i^{p}\} = \text{ViT}(\tilde{V}_i), i=1,2,...,N,
\end{equation}
where $p$ represents the number of patches in the ViT. 

\paragraph{Visual Adapter}  We utilize the global feature $v_i^{cls}$ as the feature for the i-th frame, and apply a linear layer $f(\cdot)$ to project the features of each frame into the same embedding space as that of LLM:
\begin{equation}
    z_i = f(v_i^{cls}), i=1,2,...,N.
\end{equation}
Subsequently, $Z = \{z_i\} \in \mathbb{R}^{N \times d}$ becomes the input sequence that LLM can comprehend, and $d$ is the hidden dimension of LLM.

Note that in the visual modules, we do not model the temporal relationships for the frames, inspired by the fact that the LLM itself can receive sequential input embeddings and capture their temporal relations.  

\paragraph{Input of LLM} To enable the simultaneous processing of video and text inputs, we introduce a special token, \textit{`}\textless video\textgreater\textit{'}, to represent the video content. More intuitively, if we want to ask the LLM questions about the video, we can use a mixed description, e.g., \textit{``This is a video \textless video\textgreater  can you describe this video ?''}. With this mixed description, the embedding layer will transform the textual words except for \textless video\textgreater ~ into embedding as the LLM originally does, and we will obtain a textual embedding list $[w_1, w_2, \cdots, w_M]$ for the text description, where $w_i \in R^d$ is the embedding for each word and $M$ is the word number. Then, the video feature sequence will be inserted into the embedding list, at the position of \textless video\textgreater ~ as follows,
\begin{equation}
   input=[w_1, \cdots, w_{j-1}, Z, w_{j}, \cdots, w_M ],
\end{equation}
where $j-1$ and $j$ are the indexes of the words that are close to the special token \textless video\textgreater in the original sentence. (In the previous example, $w_{j-1}$ corresponds to the word \textit{``video''}, $w_{j}$ corresponds to the word \textit{``can''}). Then LLM can further encode the input embedding list $input$ to understand the video and the user query.

\paragraph{Output for Temporal Boundaries} We employ the text format \textit{`from $s$ to $e$'} to denote a video moment, where $s$ and $e$ represent the starting and ending frame indexes of the moment, ranging from 00 to 99, with each number corresponding to a specific frame.

\subsection{Boundary-aware Training}

In contrast to the previous typical two-stage training approaches~\cite{li2023videochat, zhang2023videollama, maaz2023videochatgpt}, consisting of alignment and instruction tuning, our approach introduces an additional stage to improve the temporal understanding ability of the model. Specifically, the first stage, feature alignment, aims to train the visual adapter, to align video features with LLM's semantic space. The second stage, boundary perception, focuses on enabling LLM to develop attentional capabilities for specific moments, facilitating the understanding of various events occurring within the video. The third stage, instruction tuning, allows LLM to align with human intent and enabling more precise event localization and description. In the following sections, we will elaborate on the training methods and datasets utilized for each of these three stages.

\subsubsection{Stage 1: Feature Alignment}

\paragraph{Data Organization} In the feature alignment stage, we employ the image-text LCS-558K dataset as curated by LLaVA~\cite{liu2023llava}. This dataset is meticulously filtered to achieve a more balanced distribution of conceptual coverage. Comprising image-text pairs, we deliberately choose not to incorporate datasets containing video-text pairs with the following two considerations. Firstly, contemporary large-scale video-text datasets contain substantial textual noise, which severely impedes the alignment between visual features and textual semantics. Secondly, the transformation from visual information to text space usually suffers from information loss, e.g. when captioning an image or a video into ``a dog is running on the grass'', we may lose information about the visual details (such as the color) of the dog. Comparatively, the loss of information resulting from summarizing an image into a few words is less than that of videos. Our experiments also demonstrate the superiority of using image datasets for alignment over video dataset (a filtered subset~\cite{luo2023valley} of WebVid2M~\cite{bain2021webvid}), and even a combination of both.

\paragraph{Training Strategy} For each image-text pair \textless $I$, $T$\textgreater~ in the dataset, a special token \textless image\textgreater is directly appended before the text $T$, the embedding of this token is extracted with the visual encoder and the visual adapter as follows, denoted as $Z_I$:
\begin{equation}
    Z_I = f(\text{ViT}(I)^{cls}),
\end{equation}
and we can obtain the embedding sequence:
\begin{equation}
    input = [Z_I, w_1, w_2, \cdots, w_M].
\end{equation}
Subsequently, we can use the sequence to train the visual adapter $f$, with the original auto-regressive training objective of the LLM.

\begin{figure*}[t] 
\centering
\begin{tcolorbox}[colback=gray!20, colframe=black, text width=0.8\textwidth, title={\small Box 1: Examples of both single-turn and multi-turn QA for a video containing three events.}, fontupper=\small, fontlower=\small]
\hypertarget{box1}{}

\textbf{Single-turn QA.} \\
\textcolor{red}{$Q_1$}: Could you please detail the events that took place during different time segments in the video? \\
\textcolor{red}{$A_1$}: \textcolor{blue}{$T_1$}, from \textcolor{blue}{$s_1$} to \textcolor{blue}{$e_1$}. \textcolor{blue}{$T_2$}, from \textcolor{blue}{$s_2$} to \textcolor{blue}{$e_2$}. \textcolor{blue}{$T_3$}, from \textcolor{blue}{$s_3$} to \textcolor{blue}{$e_3$}. 
\tcblower
\textbf{Multi-turn QA.} \\
\textcolor{red}{$Q_1$}: Can you describe what occurred from \textcolor{blue}{$s_2$} to \textcolor{blue}{$e_2$} in the video? \textcolor{red}{$A_1$}: \textcolor{blue}{$T_2$}. \\
\textcolor{red}{$Q_2$}: Tell me about the events from \textcolor{blue}{$s_3$} to \textcolor{blue}{$e_3$}. \textcolor{red}{$A_2$}: \textcolor{blue}{$T_3$}. \\
\textcolor{red}{$Q_3$}: During which frames in the video can we observe \textcolor{blue}{$T_1$} happening? \textcolor{red}{$A_3$}: From \textcolor{blue}{$s_1$} to \textcolor{blue}{$e_1$}.
\end{tcolorbox}
\end{figure*}

\begin{figure*}[t] 
\centering
\begin{tcolorbox}[colback=gray!20, colframe=black, text width=0.8\textwidth, center, title={\small Box 2: The inputs to VTimeLLM in Stage 2 and Stage 3}, fontupper=\small, fontlower=\small]
\hypertarget{box2}{}
A chat between a curious user and an artificial intelligence assistant. The assistant gives helpful, detailed, and polite answers to the user's questions. \\
USER: This is a video with 100 frames: \textless video\textgreater $\backslash$n ~ \textcolor{red}{$Q_1$} ~ ASSISTANT: \textcolor{red}{$A_1$}\textless /s\textgreater \\
USER: \textcolor{red}{$Q_2$} ~ ASSISTANT: \textcolor{red}{$A_2$}\textless /s\textgreater ......
\end{tcolorbox}
\end{figure*}

\subsubsection{Stage 2: Boundary Perception}

After the training in the first stage, the LLM model becomes proficient in understanding visual information. In the second stage, we enhance the model's capabilities to comprehend sequential image frames, i.e., video, encompassing the semantic understanding of video segments while ensuring alignment with the corresponding boundaries. 

\paragraph{Data Organization} Due to the time-consuming nature of manually annotating timestamps and semantics for video segments, there is currently a lack of large-scale multi-event video-text datasets. Traditional methods align video segments with text transcripts generated by Automatic Speech Recognition (ASR). However, this approach faces challenges due to the lack of synchronicity and consistency between actions performed and spoken content, leading to weak correlations and inaccuracies in boundary annotations.

Recently, we identified the InternVid-10M-FLT~\cite{wang2023internvid} dataset, which offers a viable solution for our boundary-aware training. This dataset employs an entirely automated process to segment and annotate video clips, eliminating the need for manual intervention. Consequently, a single video may contain multiple event annotations. To ensure suitability for our study, we selected specific videos, each not exceeding 120 seconds in length. These videos encompass multiple non-overlapping event annotations, each lasting more than 3 seconds, and the average duration of these events exceeds 8\% of the video length. Thus, we curate a dataset comprising 134k videos, where each video contains multiple events and their rough temporal annotations and descriptions.

In each video, a series of events $\{s_i, e_i, T_i\}$ is contained, where $s_i$ and $e_i$ represent the start and end timestamps of a segment, ranging from 00 to 99. $T_i$ corresponds to its textual description. To transform these events into dialogue data $\{Q_1, A_1, Q_2, A_2,...\}$ suitable for training LLM, we devise two types of QA dialogues: single-turn and multi-turn, constituting 20\% and 80\% respectively. In Box \hyperlink{box1}{1}, we have provided examples of both single-turn and multi-turn QA dialogues for a video containing three events. Specifically, the task of single-turn QA is dense video captioning. $Q_1$ prompts a question requiring a comprehensive description of all events and their corresponding timestamps, while $A_1$ outputs the respective textual descriptions and timestamps in a specified format as shown in the upper box of Box \hyperlink{box1}{1}. On the other hand, multi-turn QA involves segment captioning and temporal video grounding tasks, demanding the description generation given timestamps or timestamps generation given descriptions, as shown in the lower box of Box \hyperlink{box1}{1}. In multi-turn QA, each event will be randomly queried for one of these two tasks, and the questions are not necessarily presented in the order of the events' occurrence. We design 10 templates for each task to transform events into QA dialogues, which can be found in the appendix.
 % denoted as InternVid-134K-Dense.

\paragraph{Training Strategy} We format these QA pairs according to the original LLM's format, keeping the initial system prompts intact. Moreover, we insert the statement \textit{``This is a video with 100 frames: \textless video\textgreater$\backslash$n''} before the first question. For illustration purposes, the input for VTimeLLM is presented in Box \hyperlink{box2}{2}. With the reformated sequences, we continue to employ the auto-regressive training objective, where the loss is computed exclusively on the tokens within the answer of the QA dialogues($A_1$, $A_2$, ...). To enhance training efficiency, we utilize LoRA~\cite{hu2021lora} for fine-tuning the LLM. During this stage, we keep the visual adapter $f$ frozen. Consequently, the only trainable parameters are the newly applied LoRA modules.

\subsubsection{Stage 3: Instruction Tuning}

Following the training in the second stage, our VTimeLLM model demonstrates the ability to comprehend all events within the video and align them with the corresponding timestamps. Despite the diverse templates employed, the model's output still tends to overfit the answers, which behaves more like a multi-task pretrained model while losing chatting ability with the user, e.g., when we input \textit{``What color is the coat of the man''} to the model, it may response \textit{``from 00 to 10''}. Additionally, the labels of the video-text data in the second stage are originally annotated in an automated way, which are not so accurate and noisy. To tackle the two problems, in the third stage, we incorporate high-quality dialogue data for instruction tuning, enabling the model to follow human instructions for more accurate video temporal comprehension and reasoning.

\paragraph{Data Organization} In this stage, we select a subset from ActivityNet Captions~\cite{krishna2017dense_dense45} and DiDeMo~\cite{anne2017localizing_grounding10} datasets, and transform it into a high-quality QA dialogue dataset with the assistance of Large Language Models. In contrast to InternVid which employs automated segmenting and labeling, these two datasets are entirely manually annotated, resulting in descriptions that are more detailed and temporal boundaries that are more accurate. Specifically, we carefully selected a subset of videos from the training set of ActivityNet Captions. These videos contained a minimum of three non-overlapping events, collectively covering over 90\% of the video duration, amounting to approximately 4.2k videos. Similarly, a subset of videos is being selected for the DiDeMo dataset, each containing at least two non-overlapping events and covering 40\% of the video duration. This process results in a total of about 4k videos for the DiDeMo subset. Subsequently, we also transform these videos, which contain a series of events $\{s_i, e_i, T_i\}$, into QA dialogues. However, results from the second stage of training indicate that template-based conversations lead to model overfitting. Therefore, we utilize LLM for this transformation. Specifically, we provide these events to the LLM, prompting it to assume the role of an AI visual assistant capable of analyzing the video and generate a dialogue about the video between itself and a user. The prompt can be found in the appendix. This approach results in QA dialogues that are grammatically correct, linguistically coherent, and may encompass a variety of tasks. We generate two distinct sets of dialogues for each video, yielding a final dataset comprising around 16k high-quality QA dialogues. Additionally, we observe that introducing a comparable number of other video instruction tuning datasets further enhances the descriptive capabilities of the model, with minimal impact on temporal understanding abilities. Therefore, we add an extra 20k QA pairs from the VideoInstruct100K~\cite{maaz2023videochatgpt} dataset. Overall, in this stage, a total of approximately 36k QA dialogues are used for training, which is significantly smaller than the dataset used in the second stage.

\paragraph{Training Strategy} We merge the LoRA module trained in the second stage with the original model and introduce a new LoRA module, which serves as the only trainable parameters. All other training details remain consistent with those of the second stage.
\section{Experiment}

\begin{table*}[htbp]
\small
\centering
\caption{The results of existing Video LLMs in temporal video grounding and dense video captioning tasks.}
\label{tab:main}
\setlength{\tabcolsep}{1.2mm}{
\begin{tabular}{|c|cccc|cccc|ccc|}
\hline
\multirow{3}{*}{Model} & \multicolumn{8}{c|}{Temporal Grounding}  & \multicolumn{3}{c|}{Dense Captioning}        \\ 
\cline{2-12}  
                  & \multicolumn{4}{c|}{ActivityNet} & \multicolumn{4}{c|}{Charades-STA}   & \multicolumn{3}{c|}{ActivityNet}     \\ 
                  & R@0.3 & R@0.5 & R@0.7 & mIoU & R@0.3 & R@0.5 & R@0.7 & mIoU & SODA\_c     & CIDEr     & METEOR     \\
\hline
VideoChat-7B~\cite{li2023videochat}      &  8.8   &   3.7  &  1.5   &  7.2 &  9.0   &   3.3  &   1.3  &  6.5 &   0.9          &   2.2        & 0.9            \\
VideoLLaMA-7B~\cite{zhang2023videollama}     &  6.9      &  2.1      &  0.8      &  6.5    &   10.4     &  3.8      &  0.9      &  7.1    &     1.9       &    5.8      &    1.9      \\
VideoChatGPT-7B~\cite{maaz2023videochatgpt}   &  26.4  &   13.6     &   6.1     & 18.9     &  20.0      &  7.7      &  1.7      & 13.7     &   1.9          &    5.8       &  2.1          \\
\hline
VTimeLLM-7B       & 44.0   & 27.8   & \textbf{14.3}   & 30.4 & 51.0   & 27.5   & 11.4   & 31.2 &  5.8       & \textbf{27.6}      & \textbf{6.8}        \\
VTimeLLM-13B      & \textbf{44.8}   & \textbf{29.5}   & 14.2   & \textbf{31.4} & \textbf{55.3}   & \textbf{34.3}   & \textbf{14.7}   & \textbf{34.6} &  \textbf{5.9}       & 27.2      &  6.7          \\
\hline
\end{tabular}}
\end{table*}

\subsection{Experiment Setup}

\paragraph{Tasks, Dataset, and Evaluation Metrics} To assess the capability of VTimeLLM in comprehending various event segments, we mainly conduct evaluations on two tasks: Temporal Video Grounding and Dense Video Caption. 

For the Temporal Video Grounding task, we utilize datasets from ActivityNet Captions~\cite{anne2017localizing_grounding10} and Charades-STA~\cite{gao2017tall_grounding9}. We calculate the Intersection over Union (IoU) between the time segments generated by the model and the corresponding ground truth time segments. We report mean IoU (mIoU) and recall@1, IoU$\ge m$ (R@m) metric, where $m$ values are set at $\{0.3, 0.5, 0.7\}$. 

In the case of Dense Video Captioning, we employ the ActivityNet Captions~\cite{anne2017localizing_grounding10} dataset. The evaluation process encompasses two categories of metrics. Firstly, we employ SODA\_c~\cite{fujita2020soda}, a metric specifically tailored for dense video caption tasks, taking into account the video's storyline. Secondly, we compute matched pairs between the generated events and the ground truth across IoU thresholds of $\{0.3, 0.5, 0.7, 0.9\}$, and calculate captioning metrics based on these matched pairs~\cite{yang2023vid2seq}. We report CIDEr~\cite{vedantam2015cider} and METEOR~\cite{banerjee2005meteor} averages under different IoU thresholds to provide a comprehensive analysis.

\paragraph{Implementation Details} In our study, we use Vicuna v1.5~\cite{vicuna2023} as the Large Language Model and train two versions: 7B and 13B. We use a total batch size of 128 throughout the training process. The AdamW~\cite{loshchilov2017decoupled_adamw} optimizer is applied with a cosine learning rate decay and a warm-up period. In the first training stage, the total epoch number is 1 with a learning rate of $1\times 10^{-3}$, and the subsequent second and third stages we will train for 2 epochs each with a learning rate of $1\times 10^{-4}$. The LoRA parameters are set to $r=64$ and $alpha=128$. Thanks to the efficiency of LoRA, we can complete the training of the 7B model within 30 hours with 1 RTX-4090 GPU.

\begin{table*}[htbp]
\small
\centering
\caption{Ablation study of the three-stage training strategy. In Stage 1, \textit{``}I\textit{''} and \textit{``}V\textit{''} represent the utilization of image or video datasets, and \textit{``}I+V\textit{''} signifies the merging of both datasets. In Stage 2, we compare options for Freezing or Tuning the visual adapter. In Stage 3, \textit{``}Reuse\textit{''} indicates the reuse of the LoRA from Stage 2, while \textit{``}Addition\textit{''} signifies the addition of a new LoRA module. \textit{``}\textcolor{red}{\ding{55}}\textit{''} represents the absence of training in this stage.}

\label{tab:ablation}
\setlength{\tabcolsep}{0.8mm}{
\begin{tabular}{|c|c|c|c|cccc|cccc|ccc|}
\hline
\multirow{3}{*}{Row} & \multirow{3}{*}{Stage1} & \multirow{3}{*}{Stage2} & \multirow{3}{*}{Stage3} & \multicolumn{8}{c|}{Temporal Grounding} & \multicolumn{3}{c|}{Dense Captioning}       \\ \cline{5-15} 
  &   &   &   & \multicolumn{4}{c|}{ActivityNet}                  & \multicolumn{4}{c|}{Charades-STA}  & \multicolumn{3}{c|}{ActivityNet}            \\
  &   &   &   & R@0.3 & R@0.5 & R@0.7 & mIoU & R@0.3 & R@0.5 & R@0.7 & mIoU & SODA\_c     & CIDEr     & METEOR \\ \hline
1 & \multirow{2}{*}{I}      & Freeze                  & \multirow{2}{*}{\textcolor{red}{\ding{55}}}   & 36.1  & 21.4  & 10.5  & 25.5   & 48.4  & 23.7  & 11.5  & \textbf{32.2}   & 4.6    & 17.0 & 5.4   \\ \cline{1-1} \cline{3-3}
2 &   &  Tune                &   & 35.6  & 21.4  & 10.2  & 24.8   & 48.0  & 24.4  & 12.1  & 31.6   & 4.6    & 17.0 & 5.5   \\ \cline{1-4}
3 & \multirow{2}{*}{V}      & Freeze                  & \multirow{2}{*}{\textcolor{red}{\ding{55}}}   & 33.5  & 17.9  & 8.1   & 23.9   & 46.6  & 17.7  & 6.7   & 30.5   & 4.0    & 14.1 & 5.0   \\ \cline{1-1} \cline{3-3}
4 &   &  Tune                &   & 34.8  & 18.2  & 8.5   & 24.4   & 47.9  & 19.9  & 8.0   & 31.4   & 4.2    & 14.1 & 5.2   \\ \cline{1-4}
5 & \multirow{2}{*}{I+V}    & Freeze                  & \multirow{2}{*}{\textcolor{red}{\ding{55}}}   & 31.3  & 16.5  & 6.8   & 22.2   & 47.5  & 21.9  & 9.2   & 30.6   & 4.1    & 13.4 & 5.0   \\ \cline{1-1} \cline{3-3}
6 &   &  Tune                &   & 33.5  & 17.4  & 6.9   & 23.2   & 47.1  & 21.4  & 8.6   & 30.6   & 4.0    & 13.6 & 4.8   \\ \cline{1-4}
7 & \textcolor{red}{\ding{55}} &  Tune                & \textcolor{red}{\ding{55}} & 42.2  & 22.7  & 11.5  & 29.8   & 40.0  & 4.9   & 0.0   & 27.5   & 3.7    & 10.2 & 5.0   \\ \cline{1-4}
8 & I & \textcolor{red}{\ding{55}} & Addition & 31.0  & 18.1  & 7.7   & 22.9   & 37.5  & 21.8  & 6.1   & 22.8   & 4.2    & 16.0 & 5.0   \\ \cline{1-4}
9 & I & Freeze                  & Reuse                  & 39.3  & 26.6  & 13.0  & 28.1   & 49.7  & \textbf{29.8}  & \textbf{13.3}  & 30.9   & 5.2    & 23.2 & 6.1   \\ \cline{1-4}
10                   & I & Freeze                  & Addition                   & \textbf{44.0}  & \textbf{27.8}  & \textbf{14.3}  & \textbf{30.4}   & \textbf{51.0}  & 27.5  & 11.4  & 31.2   & \textbf{5.8}    & \textbf{27.6} & \textbf{6.8}      \\ \hline
\end{tabular}}
\end{table*}

\subsection{Main Results}

We evaluate the capabilities of existing Video LLMs in temporal video grounding and dense video captioning tasks, as shown in Table 1. Detailed information about the evaluation process can be found in the appendix. VTimeLLM-7B outperforms these Video LLMs of the same size by a significant margin. Upon further scaling up the model to 13B parameters, we observe minor changes in performance on ActivityNet tasks, while the temporal grounding ability improves on Charades-STA. It is worth mentioning that our training dataset does not include Charades-STA training data, indicating that increasing the scale of our VTimeLLM model enhances its out-of-distribution generalization capability.

We provide several possible explanations to account for the poor performance of other models: firstly, both VideoChat and VideoLLaMA extract only N=8 frames as input, making it challenging for them to achieve a fine-grained understanding of the video content. Secondly, the commonly used LLM (Vicuna) lacks robust positional awareness in input sequences. For instance, when posed with the question \textit{``What is the position of the word `video' in the phrase `a video clip' ?''}, it may erroneously respond, \textit{``The word `video' appears at position 67.''} Relying solely on a limited set of temporally annotated data for instruction tuning is insufficient to address this issue. Therefore, it is essential to integrate boundary-aware training to achieve precise video comprehension.

\begin{table*}[htbp]
\small
\centering
\caption{The results of video dialogue on video-based generative performance benchmarking.}
\label{tab:benchmark}
\setlength{\tabcolsep}{1.4mm}{
\begin{tabular}{|l|c|c|c|c|c|c|}
\hline

Evaluation Aspect                                      & VideoLLaMA & LLaMA-Adapter & VideoChat & VideoChatGPT& BT-Adapter & \textbf{VTimeLLM}     \\ \hline
Temporal Understanding                                 & 1.82       & 1.98          & 1.94      & 1.98         & 2.34       & \textbf{2.49} \\
Correctness of Information                             & 1.96       & 2.03          & 2.23      & 2.40         & 2.68       & \textbf{2.78} \\
Detail Orientation     & 2.18       & 2.32          & 2.50      & 2.52         & 2.69       & \textbf{3.10} \\
Contextual Understanding & 2.16       & 2.30          & 2.53      & 2.62         & 3.27       & \textbf{3.40} \\
Consistency           & 1.79       & 2.15          & 2.24      & 2.37         & 2.46       & \textbf{2.47} \\
Mean                                                   & 1.98       & 2.16          & 2.29      & 2.38         & 2.69       & \textbf{2.85} \\ \hline
\end{tabular}
}
\end{table*}

\subsection{Ablation Study}

In this section, we provide detailed ablations about our three-stage training strategy through experiments on the 7B model, as illustrated in Table \ref{tab:ablation}. In the ablation, our most conerned questions and their results are provided in the following. 

\paragraph{Q1: How to train a good visual adapter?}  In contrast to other Video LLMs, we utilize a pure image modality for the first stage and find it to be superior across all metrics than using a pure video modality (Rows 1, 2 vs Rows 3, 4). This effectiveness of using images to alignment could be attributed to the higher quality and reduced information loss in image datasets. Additionally, using pure images outperforms the fusion of two modal datasets (Rows 1, 2 vs Rows 5, 6). This could be due to the significant disparity in tasks, where describing a single frame event and describing a sequence of 100 frames events pose distinct challenges for model fitting. 

Another question arises during the following stage: should the previously pretrained visual adapter be tuned or frozen? Upon comparing Row 1 vs Row 2, Row 3 vs Row 4, and Row 5 vs Row 6, we observed minor difference in performance between the two approaches. To retain the comprehensive information acquired during the pretraining stage, we opt to freeze the parameters of the visual adapter in the latter two stages.

\paragraph{Q2: Should the LoRA from stage 2 be reused in stage 3?}  Upon comparing Row 9 to Row 10, it is evident that in stage 3, merging the LoRA module from the second stage with the LLM parameters and additionally incorporating another LoRA module yields superior results. This approach ensures that the temporal understanding capabilities acquired during stage 2 are effectively preserved within the model.
 
\paragraph{Q3: Is every training stage necessary?} By comparing Rows 1\textasciitilde 6 with Row 7, we observe a substantial disparity in the model's performance on temporal grounding when training without stage 1. The scores are abnormally high on the ActivityNet dataset, while significantly low on the Charades-STA dataset. Upon careful analysis of the outputs under this setting, we find that the model has not effectively learned to localize events. Instead, it
tends to predict a temporal segment spanning nearly the entire video (e.g., from 00 to 95). In such cases, if the ratio of ground truth duration to video length is denoted as $x$, the IoU with the model's output is approximately $x$. The ActivityNet dataset contains a significant number of long samples, with 20\% of queries having $x>0.5$, leading to an inflated evaluation metric. Conversely, in the Charades-STA dataset, $x$ rarely exceeds 0.5, demanding more precise localization~\cite{lan2023closer}. However, the model without stage 1 training fails to achieve it. Moreover, the model's performance in dense captioning tasks is unsatisfactory, which also highlights the essential nature of the feature alignment stage.

The necessity of stage 2 can be demonstrated by comparing Row 8 with Rows 9, 10. Despite the higher quality of annotations in stage 3, the limited dataset size hinders the model's ability to achieve a robust temporal understanding through stage 3 training alone. Models trained solely in stage 3 exhibit inferior performance across various tasks compared to those that have undergone preliminary training in stage 2.

After stage 3 training, the model exhibits comprehensive improvement in the tasks outlined in the table (Row 1 vs Row 10). Furthermore, it regains chatting ability, enabling it to respond to a wide range of questions posed by humans.

\subsection{Video Dialogue Performance}
% Video-Based Text Generation Performance

Besides the ability for fine-grained video understanding tasks, we explore whether VTimeLLM can address a broader range of questions through dialogue. We employ the Video-ChatGPT~\cite{maaz2023videochatgpt} benchmark and conduct an evaluation of video-based generative performance. This benchmark covers many questions associated with five key aspects. GPT-3.5 assigns a score, not exceeding 5, to the model-predicted answer based on the question and the correct answer. We present the average scores in Table \ref{tab:benchmark} and compare VTimeLLM with all existing Video LLMs, including VideoLLaMA~\cite{zhang2023videollama}, LLaMA-Adapter~\cite{zhang2023llama-adapter}, VideoChat~\cite{li2023videochat}, VideoChatGPT~\cite{maaz2023videochatgpt} and BT-Adapter~\cite{liu2023one_btadapter}.

Thanks to the fine-grained video comprehension capabilities, VTimeLLM achieves state-of-the-art results in all aspects. The most substantial improvement is observed in the aspect of detail orientation, where VTimeLLM achieves a noteworthy enhancement of +0.41 (15.2\%). We attribute this progress to two primary factors. Firstly, the image-based training in stage 1 ensures comprehensive preservation of visual details in individual frames, facilitating a detailed understanding of spatial dimension. Secondly, the temporal-aware training employed in the second and third stages enables VTimeLLM to capture multiple events within videos, enhancing its ability to depict details of temporal dimension.

\begin{figure}[htbp]
  \centering
  \includegraphics[width=.9\columnwidth]{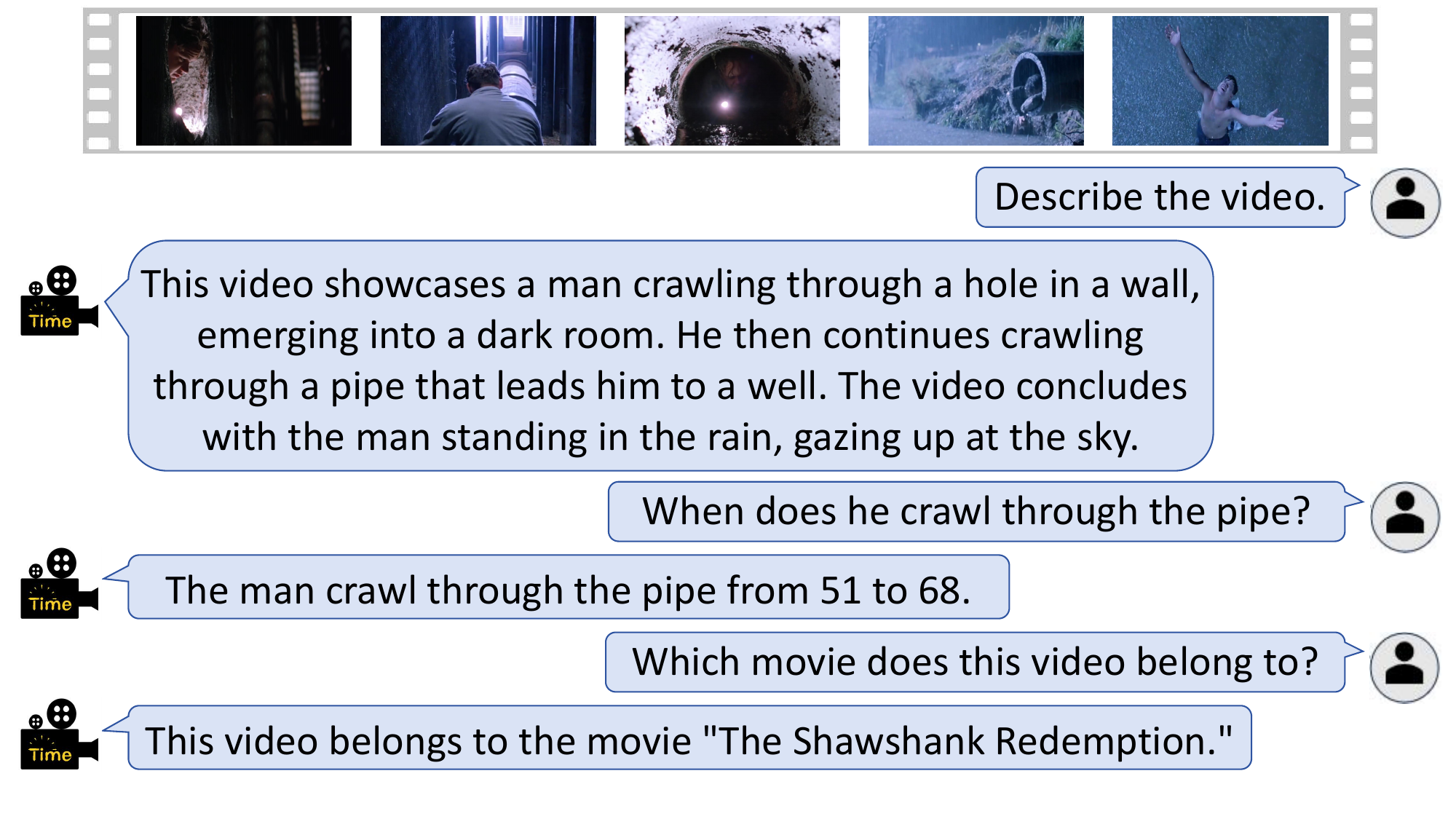}
  \caption{A qualitative example of video dialogue. The video is 160 seconds long.}
  \label{fig:example}
\end{figure}

To better illustrate the video dialogue performance of VTimeLLM, we present a qualitative example, as shown in Figure \ref{fig:example}.

\section{Conclusion}

In this work, we introduce VTimeLLM, a Video LLM capable of comprehending multiple events within a video and providing precise temporal boundaries. We unify video tasks demanding fine-grained comprehension, such as temporal video grounding and dense video captioning, and pioneer their addressing using Video LLM. Specifically, we propose a three-stage temporal-aware training framework. This framework utilizes large-scale image-text data for feature alignment, leverages extensive multi-event video-text data along with temporal-related question-answering to enhance temporal awareness, and employs instruction tuning on a high-quality dialogue dataset to improve temporal reasoning ability. Extensive experiments demonstrate that VTimeLLM outperforms existing Video LLMs significantly across various tasks, particularly excelling in fine-grained temporal-related video tasks, showing VTimeLLM's superior ability for video understanding and reasoning.

{
    \small
    \bibliographystyle{ieeenat_fullname}
    \bibliography{arxiv}
}

\clearpage
\setcounter{page}{1}
% \maketitlesupplementary
\appendix
\section{More Examples}

\begin{figure}[htbp]
  \centering
  \includegraphics[width=\columnwidth]{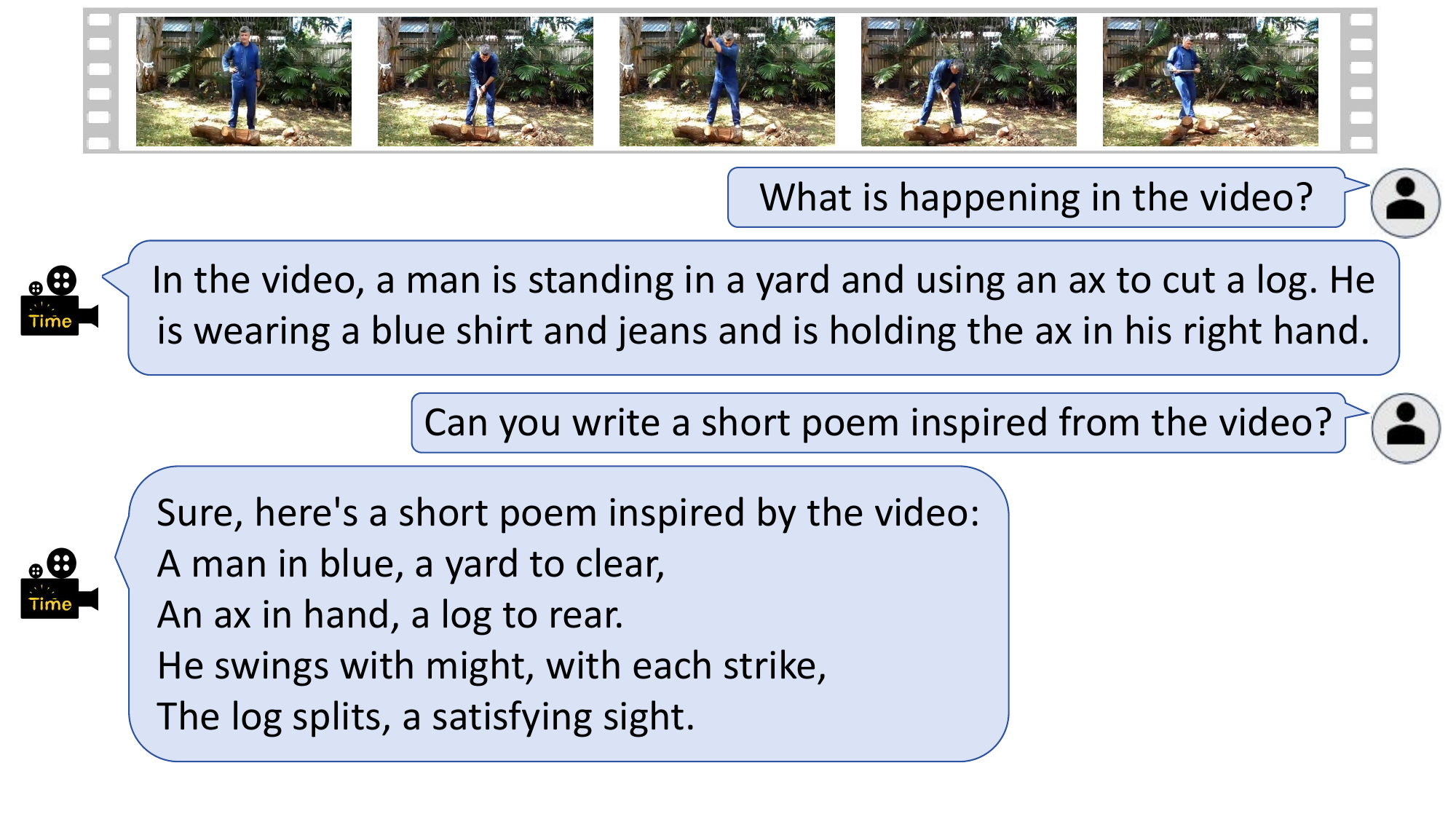}
  \caption{Creative Task. Our VTimeLLM comprehends the visual information and crafts a poem inspired by it.}
  \label{fig:ex1}
\end{figure}

\begin{figure}[htbp]
  \centering
  \includegraphics[width=\columnwidth]{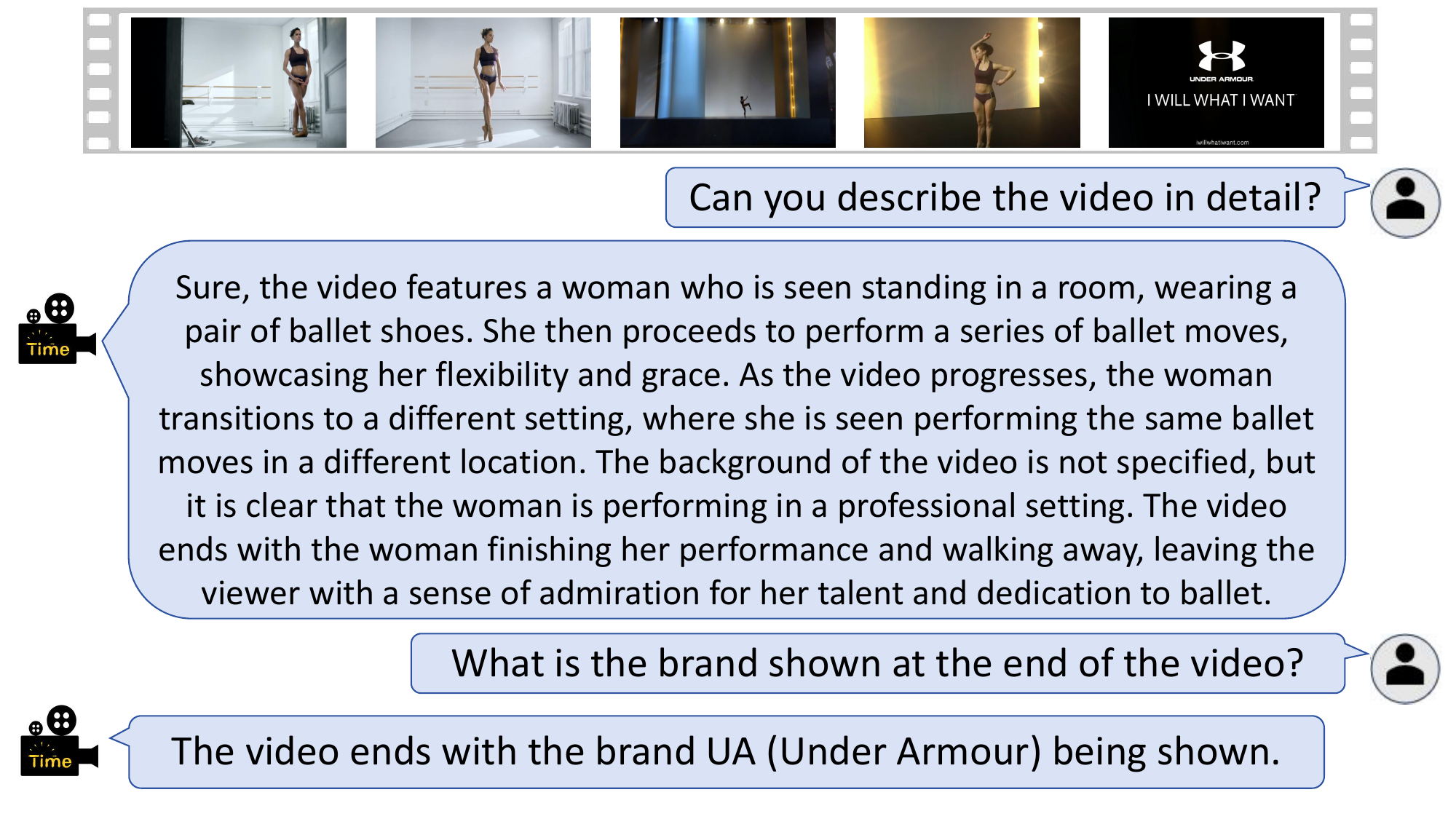}
  \caption{Fine-grained Understanding Task. Our VTimeLLM comprehends multiple events within the video, as well as the specific visual content within individual events.}
  \label{fig:ex2}
\end{figure}

\begin{figure}[htbp]
  \centering
  \includegraphics[width=\columnwidth]{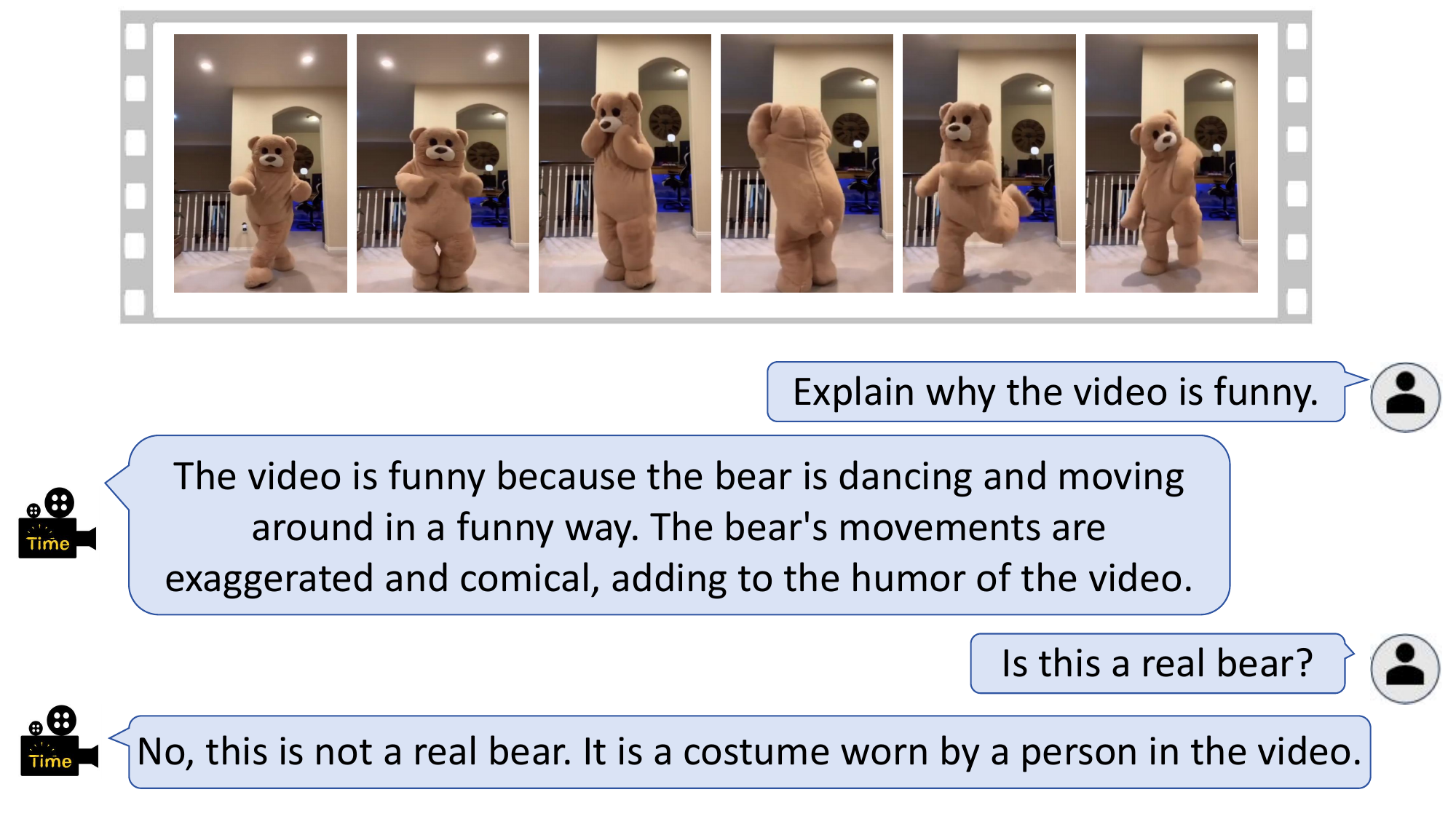}
  \caption{Video Reasoning Task. our VTimeLLM responds to several questions requiring reasoning based on a comprehensive understanding of visual content.}
  \label{fig:ex3}
\end{figure}

We showcase additional examples of video dialogues across various tasks, encompassing a creative task (Figure \ref{fig:ex1}), a fine-grained understanding task (Figure \ref{fig:ex2}), and a video reasoning task (Figure \ref{fig:ex3}). In the creative task (Figure \ref{fig:ex1}), our VTimeLLM demonstrates a remarkable capacity to comprehend visual information and subsequently craft a poem inspired by it. This achievement is attributed to we freeze the LLM at all three stages of training, thereby preserving its ability for engaging in creative dialogue. In the fine-grained understanding task (Figure \ref{fig:ex2}), our VTimeLLM comprehends multiple events within the video, as well as the specific visual content within individual events. This demonstration underscores its proficiency in grasping temporal and spatial details, a capability attributed to our three-stage training strategy. In the video reasoning task (Figure \ref{fig:ex3}), our VTimeLLM responds to several questions requiring inference, showing its capacity to engage in reasoning based on a comprehensive understanding of visual content.

\section{Templates and Prompts}

In Stage 2, we need to transform events $\{s_i, e_i, T_i\}$ into template-based QA, where $s_i$ and $e_i$ represent the start and end timestamps of a segment, ranging from 00 to 99. $T_i$ corresponds to its textual description.  For a given sequence of events, there is a 20\% probability of transformation into single-turn QA, completing a dense caption task where all events are described within a single answer. Conversely, there is an 80\% probability of transformation into Multi-turn QA. In this scenario, each event is individually queried and answered within a dialogue, in the form of two tasks, event captioning or temporal grounding. We provide 10 templates for each task, as shown in Box \hyperlink{box3}{3}.

In Stage 3, we need to transform events into high-quality dialogue. This is accomplished by providing a prompt to a text-based LLM(Vicuna-7B v1.5). The prompt can be found in Box \hyperlink{box4}{4}. In the prompt, specific timestamps are not provided because their inclusion does not enhance the LLM's comprehension of temporal relationships. On the contrary, they may introduce errors into the dialogue. Consequently, events are presented in a sequential order, accompanied by specific symbols e.g., \textless s1\textgreater, \textless e1\textgreater in the box, denoting the timestamps. The generated dialogue is expected to integrate temporal perception and reasoning.

\section{Evaluation Process}

In this section, we provide a detailed process on the evaluation of temporal grounding and dense captioning tasks for VTimeLLM and other Video LLMs.

\subsection{Evaluation of VTimeLLM}

For VTimeLLM that has undergone only stages 1, 2 training without stage 3, the input and output formats remain entirely consistent with the template. Consequently, we can directly employ the templates in Box \hyperlink{box3}{3} as queries. Specifically, for the dense captioning task, we employ $Q_{D1}$, i.e., \textit{``Could you please detail the events that took place during different time segments in the video?''} as the query. For the temporal grounding task, we employ queries $Q_{T1}$, $Q_{T2}$, and $Q_{T3}$ to compute IoU for their respective outputs, and we report the average metrics. The performance obtained from different queries is similar.

VTimeLLM that has undergone stage 3 training demonstrate commendable instruction-following ability, and the performance may vary with different queries. For example, the inclusion of the phrase \textit{``in detail''} in the query leads to a more detailed description of the video. For the dense captioning task, we utilize the following query: \textit{``Could you please describe the events in the video in detail? Be specific about the activities of individuals, their surroundings, and interactions with others. The output should be in JSON format, structured as follows: \{`event': `xx', `timestamps': `from xx to xx'\}.''} We find that this query outperforms $Q_{D1}$ across various metrics by approximately 10\%. For the temporal grounding task, we continue to report the average results of queries $Q_{T1}$, $Q_{T2}$, and $Q_{T3}$, with metrics for each query remaining consistently close. Notably, even with the adoption of a simpler query such as \textit{`When does $T_i$ happen?''}, we achieve comparable results, underscoring the stability of outputs in this task.

\subsection{Evaluation of other Video LLMs}

For other Video LLMs (VideoLLaMA, VideoChat, and VideoChatGPT) that we test in our study, we try our best to assess their optimal performance as they were not trained on these tasks. Our testing methodology follows several principles: First, we include video duration $D$ in the query. Second, as these models often fail to adhere to our prompt for outputting in JSON format, we apply multiple regular expressions to format the output. This successfully handles over 70\% of the outputs. For these outputs cannot be processed, we exclude the corresponding data from metric calculations. Third, we design multiple queries and select the one yielding the best performance as the final result. For example, in our experiment, we find that the best query for VideoChatGPT in the dense captioning cask is: \textit{``This video has a duration of $D$ seconds. From which second to which second in the video, what event happens? Be specific about the activities of individuals, their surroundings, and interactions with others. List the events in the format: 1. From x1 second to y1 second: event1. $\backslash$n ~ 2. From x2 second to y2 second: event2.$\backslash$n ~ ...''}

\begin{figure*}[t] % 使用figure*环境创建双栏的浮动体
\centering

\begin{tcolorbox}[colback=gray!20, colframe=black, text width=0.9\textwidth, title={Box 3: Templates to transform events $\{s_i, e_i, T_i\}$ into QA dialogues, 10 templates for each task.}]
\hypertarget{box3}{}

\textbf{Dense Captioning (task of Single-turn QA):}

$Q_{D1}$: Could you please detail the events that took place during different time segments in the video? \\ 
$Q_{D2}$: I'm curious about what happened at different points in the video. Could you please describe the events? \\ 
$Q_{D3}$: Could you provide a summary of the incidents that occurred at various timestamps in the video? \\ 
$Q_{D4}$: I'd like to know what events transpired during specific time intervals in the video. Could you please elaborate? \\ 
$Q_{D5}$: Can you give me a breakdown of the occurrences at different time stamps in the video? \\ 
$Q_{D6}$: I'm interested in understanding the events that unfolded at different points in the video. Could you please specify? \\ 
$Q_{D7}$: Could you outline the incidents that happened during different time periods in the video? \\ 
$Q_{D8}$: I'm trying to grasp the sequence of events in the video. Could you please outline what happened at different times? \\ 
$Q_{D9}$: Can you go through the video and describe what took place at different time intervals? \\ 
$Q_{D10}$: I'd appreciate it if you could provide a detailed account of the events that occurred at different timestamps in the video. \\
$A_{D}$: \textcolor{blue}{$T_1$}, from \textcolor{blue}{$s_1$} to \textcolor{blue}{$e_1$}. \textcolor{blue}{$T_2$}, from \textcolor{blue}{$s_2$} to \textcolor{blue}{$e_2$}. \textcolor{blue}{$T_3$}, from \textcolor{blue}{$s_3$} to \textcolor{blue}{$e_3$}...... \\

\tcblower

\textbf{Event Captioning (One task in Multi-turn QA):}

$Q_{E1}$: Can you describe what occurred from \textcolor{blue}{$s_i$} to \textcolor{blue}{$e_i$} in the video? \\ 
$Q_{E2}$: Could you tell me what happened from \textcolor{blue}{$s_i$} to \textcolor{blue}{$e_i$} in the video? \\ 
$Q_{E3}$: What transpired from \textcolor{blue}{$s_i$} to \textcolor{blue}{$e_i$} in the video? \\ 
$Q_{E4}$: Describe what took place from \textcolor{blue}{$s_i$} to \textcolor{blue}{$e_i$} in the video. \\ 
$Q_{E5}$: Tell me about the events from \textcolor{blue}{$s_i$} to \textcolor{blue}{$e_i$} in the video. \\ 
$Q_{E6}$: What was going on from \textcolor{blue}{$s_i$} to \textcolor{blue}{$e_i$} in the video? \\ 
$Q_{E7}$: Please recount what occurred from \textcolor{blue}{$s_i$} to \textcolor{blue}{$e_i$} in the video. \\ 
$Q_{E8}$: Explain what happened from \textcolor{blue}{$s_i$} to \textcolor{blue}{$e_i$} in the video. \\ 
$Q_{E9}$: Provide details about the events from \textcolor{blue}{$s_i$} to \textcolor{blue}{$e_i$} in the video. \\ 
$Q_{E10}$: Share what transpired from \textcolor{blue}{$s_i$} to \textcolor{blue}{$e_i$} in the video. \\
$A_{E}$: \textcolor{blue}{$T_i$}. \\

\textbf{Temporal Grounding (One task in Multi-turn QA):}

$Q_{T1}$: During which frames can we see \textcolor{blue}{$T_i$} happening in the video? \\ 
$Q_{T2}$: Between which frames is \textcolor{blue}{$T_i$} visible in the video? \\ 
$Q_{T3}$: At what point in the video can we observe \textcolor{blue}{$T_i$} taking place? \\ 
$Q_{T4}$: Between which two frames can we witness \textcolor{blue}{$T_i$} occurring in the video? \\ 
$Q_{T5}$: During which frames in the video can we observe \textcolor{blue}{$T_i$} happening? \\ 
$Q_{T6}$: At which time interval in the video can we see \textcolor{blue}{$T_i$} occurring? \\ 
$Q_{T7}$: Between which frames can we find \textcolor{blue}{$T_i$} taking place in the video? \\ 
$Q_{T8}$: At what point in the video can we witness \textcolor{blue}{$T_i$} happening? \\ 
$Q_{T9}$: Between which two frames in the video can we observe \textcolor{blue}{$T_i$} taking place? \\ 
$Q_{T10}$: During which frames does \textcolor{blue}{$T_i$} occur in the video? \\
$A_{T}$: From \textcolor{blue}{$s_i$} to \textcolor{blue}{$e_i$}. \\

\end{tcolorbox}

\end{figure*}
\begin{figure*}[t] % 使用figure*环境创建双栏的浮动体
\centering

\begin{tcolorbox}[colback=gray!20, colframe=black, text width=0.9\textwidth, title={Box 4: Prompt for LLM in Stage 3, designed to transform a series of events into high-quality dialogue.}]
\hypertarget{box4}{}

You are an AI visual assistant with the task of analyzing a single video.\\Craft a conversation between yourself and a user discussing the video's content. Develop responses that embody the persona of an active visual AI assistant, capable of observing the video and providing insightful answers. Include inquiries about temporal perception and reasoning, like events preceding or succeeding specific occurrences, or requesting timestamps for particular actions or events.\\Ensure that the questions can be definitively answered based on the observable video content or confidently ascertainable absence from the video. Utilize the timestamps \textless s?\textgreater  and \textless t?\textgreater  to create contextual questions considering the temporal relationships between events. The conversations should be concise.\\\\Here's an illustrative example:\\=== example start ===\\Events:\\from \textless s1\textgreater  to \textless e1\textgreater : A man and woman play rock paper scissors, the woman wins and smiles.\\from \textless s2\textgreater  to \textless e2\textgreater : The woman puts a blindfold on.\\from \textless s3\textgreater  to \textless e3\textgreater : The woman continues playing rock-paper-scissors with the man and wins again.\\from \textless s4\textgreater  to \textless e4\textgreater : The woman gives the man a hug.\\\\Dialogue:\\User: Could you provide a brief overview of the video's content?\\Assistant: Certainly! In the video, a man and a woman engage in a game of rock-paper-scissors. The woman emerges victorious and shares a smile. Subsequently, she places a blindfold on. She then proceeds to win another round of rock-paper-scissors against the man. The video concludes with the woman embracing the man warmly.\\User: Can you pinpoint when the woman achieved victory in the game twice?\\Assistant: Certainly. The first victory occurs from \textless s1\textgreater  to \textless e1\textgreater , while the second triumph takes place from \textless s3\textgreater  to \textless e3\textgreater .\\User: I'm curious about the interaction between \textless s4\textgreater  and \textless t4\textgreater . Could you elaborate?\\Assistant: Absolutely. During the interval from \textless s4\textgreater  to \textless t4\textgreater , the woman conveys her emotions through a heartfelt embrace, demonstrating her genuine affection for the man.\\User: What might be the underlying reason for the woman's affectionate hug?\\Assistant: The woman's affectionate hug likely stems from her desire to uplift the man's spirits after his loss in the rock-paper-scissors game.\\=== example end ===\\\\Events:\\ 
from \textless s1\textgreater  to \textless e1\textgreater : \textcolor{blue}{$T_1$}.\\
from \textless s2\textgreater  to \textless e2\textgreater : \textcolor{blue}{$T_2$}.\\
from \textless s3\textgreater  to \textless e3\textgreater : \textcolor{blue}{$T_3$}.\\
...... \\\\
Dialogue:

\end{tcolorbox}

\end{figure*}

\end{document}